# Semi-supervised GAN-based Segmentation of CTPA Scans for Pulmonary Emboli Boundary Extraction

Maria Thoma
Dept. of Computer Science and Biomedical Informatics
University of Thessaly
Lamia, Greece
thmaria@uth.gr

Michalis A. Savelonas
Dept. of Computer Science and Biomedical Informatics
University of Thessaly
Lamia, Greece
msavelonas@uth.gr

Foteini Malli
Dept. of Nursing
University of Thessaly
Larissa, Greece
fmalli@uth.gr

Dimitris K. Iakovidis
Dept. of Computer Science and Biomedical Informatics
University of Thessaly
Lamia, Greece
diakovidis@uth.gr

*Abstract*— **Pulmonary embolism (PE) is a pulmonary vascular disease associated with high morbidity and mortality, where delayed diagnosis can be fatal. While computed tomography pulmonary angiography (CTPA) is the clinical gold standard for PE detection, interpreting these scans manually is time-consuming and error-prone. In this work, we propose a semi-supervised segmentation method, employing generative adversarial networks (GANs) to leverage large pools of unlabeled CTPA scans. The proposed method addresses more effectively the challenges of CTPA image segmentation than previous methods by employing the dice loss and the cross-entropy loss to tackle the highly imbalanced nature of CTPA scans, the feature matching loss to encourage the generator to produce anatomically plausible segmentation maps that are indistinguishable from ground truth, as well as the self-training loss to reinforce learning from highly-confident pseudo-labeled samples. The experimental evaluation on a publicly available dataset shows that the proposed method maintains accurate segmentation, despite a limited number of labeled CTPA scans. with a Dice coefficient of 88.75% and a 95th percentile Hausdorff distance (HD95) of 1.74, outperforming fully supervised baselines.**

## I. Introduction

Pulmonary Embolism (PE) is a potentially life-threatening condition caused by the sudden blockage of one or more pulmonary arteries, typically due to blood clots that travel from the deep veins of the legs or other parts of the body [1]. PE can lead to impaired gas exchange, increased pulmonary artery pressure, and right heart strain, often resulting in severe clinical consequences if not diagnosed and treated promptly. Early and accurate detection is critical, as timely anticoagulation or thrombolytic therapy can dramatically reduce mortality [1]. However, the clinical presentation of PE is often nonspecific - symptoms like shortness of breath, chest pain, and tachycardia may overlap with several cardiopulmonary conditions - making imaging an essential element of the diagnostic workflow [2].

Computed Tomography Pulmonary Angiography (CTPA) has emerged as the imaging modality of choice for the diagnosis of PE, due to its wide availability, rapid acquisition time, and high spatial resolution [1], [2]. By providing detailed visualization of the pulmonary arteries, following the administration of contrast material, CTPA allows radiologists to directly observe intraluminal filling defects indicative of emboli. In addition to confirming or ruling out PE, CTPA can also reveal alternative or concomitant pathologies, such as pneumonia, pleural effusions, or right heart strain. Despite its diagnostic value, the interpretation of CTPA scans can be time-consuming and subject to inter-observer variability, especially in emergency settings or when emboli are subtle or located in peripheral vessels [3]. In this light, computational methods could potentially accelerate diagnosis and strengthen objectivity.

Automated analysis and segmentation of CTPA scans present several technical challenges. The pulmonary vasculature is complex and highly variable across individuals, and emboli can appear in a wide range of locations, sizes, and contrast levels. Small or subsegmental emboli often occupy only a tiny fraction of the image and may be obscured by artifacts, motion blur, or anatomical noise from overlapping structures [4]. Moreover, the annotation of PE regions for supervised training is labor-intensive and prone to subjectivity, limiting the availability of large, high-quality labeled datasets.

Prior to the rise of the deep learning paradigm, medical image segmentation relied heavily on traditional techniques such as thresholding, region growing, level-sets, and model-based approaches like graph cuts [5]. These methods typically required handcrafted features and carefully tuned parameters to capture anatomical boundaries and detect abnormalities. While often effective in well-controlled settings, their performance degraded significantly in the presence of noise, low contrast, or anatomical variability - common issues in

This work is part of the European project SEARCH, which is supported by the Innovative Health Initiative Joint Undertaking (IHI JU) under grant agreement No. 101172997. The JU receives support from the European Union's Horizon Europe research and innovation programme and COCIR, EFPIA, Europa Bio, MedTech Europe, Vaccines Europe, Medical Values GmbH, Corsano Health BV, Syntheticus AG, Maggioli SpA, Motilent Ltd, Ubitech Ltd, Hemex Benelux, Hellenic Healthcare Group, German Oncology Center, Byte Solutions Unlimited, AdaptIT GmbH. Views and opinions expressed are however those of the author(s) only and do not necessarily reflect those of the aforementioned parties. Neither of the aforementioned parties can be held responsible for them.

CTPA scans. Moreover, their reliance on local intensity cues and limited context made it difficult to segment subtle emboli or differentiate them from overlapping vascular or parenchymal structures [5]. These limitations spurred the search for data-driven, robust techniques capable of capturing complex spatial patterns and semantic features.

The advent of Convolutional Neural Networks (CNNs) led to significant advances in medical image segmentation. Architectures such as U-Net and its variants became widely adopted, due to their ability to learn hierarchical representations from data and integrate multiscale context through skip connections [6]. Despite their success, CNNs exhibit certain limitations when applied to CTPA image segmentation: their inductive biases (e.g., locality and translation invariance) can hinder long-range dependency modeling, and their performance often drops when training data is limited or exhibits domain shifts. More recently, Vision Transformers (ViTs) have been introduced to address these issues by modeling global self-attention, offering promising results in medical image analysis. However, ViTs generally require large datasets and extensive computational resources, whereas they may struggle to retain fine-grained spatial details, which are critical for small emboli detection [7],[8]. In this context, Generative Adversarial Networks (GANs) offer a compelling alternative by framing segmentation as a structured prediction task, enhanced by adversarial learning. The latter encourages the model to produce anatomically plausible and sharper segmentations, even under limited supervision or weak signal conditions [9].

The majority of previous works has focused on supervised CTPA image segmentation, which relies on the availability of detailed pixel-wise annotations. For instance, Ajmera et al. [10] proposed a deep learning method based on U-Net to extract features that were co-evaluated for scan-level PE classification. Similarly, Liu et al. [11] introduced CAM-WNet, a two-stage U-Net-based model that incorporates Coordinate Attention and Pyramid Pooling Modules to enhance PE segmentation. More recently, Chen et al. [12] proposed SCUNet++, which employs a SWIN Transformer encoder, combined with CNN bottlenecks, while Nitha et al. [13] employed a fine-tuned Attention U-Net.

Although supervised methods have demonstrated strong performance, they rely heavily on the availability of large, annotated datasets. However, acquiring pixel-level annotations in medical imaging - especially in the case of PE annotations in CTPA images - is both time-consuming and labor-intensive. This highlights the need for methods that maintain high segmentation performance despite using limited labeled data for training, such as semi-supervised, weakly supervised or even fully unsupervised methods.

Despite the success of semi-supervised and weakly-supervised methods in natural image segmentation [14]- [15], their application to CTPA image segmentation for PE boundary extraction remains limited. For example, Pu et al. [16] proposed a fully unsupervised method that performs CTPA image segmentation and PE boundary extraction using a five-stage pipeline, including artery–vein segmentation, vessel diameter partitioning, and contrast-based detection of low-density regions. High-confidence regions are then used as pseudo-labels to train a 3D CNN. While this approach avoids manual annotations, it often struggles to detect small or low-contrast emboli due to their subtle appearance and ambiguous boundaries. More recently, Cheng et al. [17] applied a semi-supervised GAN-based framework. Their method utilizes a small proportion of labeled data, with an HRNet-based generator, designed to extract high-resolution features for segmentation. The discriminator, trained on both predicted masks and feature representations, is tasked with distinguishing between ground truth and predicted confidence maps, while also maximizing their similarity. However, segmentation results have limited accuracy, with a Dice score of 0.56.

In this work, we propose a semi-supervised adversarial learning method for PE boundary extraction in CTPA scans, for which pixel-wise annotations are scarce and expensive to obtain. The proposed method is designed to leverage both a set of labeled CTPA scans, as well as a larger pool of unlabeled scans by employing semi-supervised learning in the form of adversarial training. The latter progressively incorporates high-confidence predictions, as pseudo-labels on unlabeled data. The proposed method addresses more effectively the challenges of CTPA image segmentation than previous methods by employing a hybrid Dice loss and cross-entropy loss to tackle the highly imbalanced nature of CTPA scans and the small size of some PEs, feature matching loss to guide the generator towards anatomically plausible segmentation maps that are indistinguishable from ground truth, and self-training loss to reinforce learning from highly-confident pseudo-labeled samples.

The rest of this manuscript is structured as follows: Section II presents the proposed semi-supervised adversarial learning framework, Section III presents and discusses experimental results, and Section IV summarizes the main conclusions of this work.

## II. Semi-supervised Adversarial Learning

The proposed method employs a conditional GAN [18], based on s4GAN [19]. Although this conditional GAN was originally developed for natural image segmentation, in the context of the proposed method it is employed to generate anatomically plausible artificial ground truth masks for a large amount of unlabeled CTPA scans. These artificial masks are used for training, along with a small number of manually labelled CTPA scans. The proposed method encompasses: i) SCUNet++ [12] for segmentation, considering its performance in supervised CTPA image segmentation for PE detection, and ii) a hybrid loss function that combines cross-entropy loss and Dice loss [20], in order to aid the network to focus both on capturing small-sized regions of PE and on improving pixel-level accuracy, iii) a feature matching loss function to encourage the generator to produce anatomically plausible segmentation maps, iv) and a self-training loss function to reinforce learning from highly-confident pseudo-labeled samples.

The s4GAN model consists of two main components: the generator and the discriminator. The generator operates as a segmentation network. It uses unlabeled images for training, as well as images labeled with ground truth masks, and generates artificial ground truth masks for the unlabeled images. The discriminator receives as input a concatenation of the input image and its corresponding segmentation mask - either a real ground-truth mask (for labeled data) or an artificial mask (for unlabeled data). It outputs a probability indicating whether the input pair is real (from labeled data) or artificial (from the generator). These two components are trained in an adversarial manner, guiding the generator towards realistic and accurate artificial ground truth masks. Figure 1 provides an overview of the proposed method.

### *A. Generator*

The generator, which serves as the segmentation network, is trained to 'fool' the discriminator, by generating artificial ground truth masks that closely resemble real ground truth masks. It is optimized by means of a combination of supervised loss terms: the Dice loss and cross-entropy loss, as well as the feature matching and self-training losses.

Dice loss and cross-entropy loss address the highly imbalanced nature of CTPA scans of PE at the image level, where PE occupy only a tiny fraction of the image. The Dice loss quantifies the overlap between the predicted region and the ground truth region, and is particularly effective in handling imbalanced class distributions. It is defined as follows:

$$L_{\mathrm{Dice}} = 1 - DSC, \tag{1}$$

where DSC refers to the dice coefficient provided by:

$$DSC = \frac{2\sum_{i=1}^{N} p_i g_i}{\sum_{i=1}^{N} p_i^2 + \sum_{i=1}^{N} g_i^2} \tag{2}$$

where $N$ is the number of pixels, $p_i$ is the probability that the $i$-th pixel belongs to the target class, and $g_i$ is the ground truth label of the $i$-th pixel.

One of the main drawbacks of Dice loss is that it can become unstable during training, when the target region is small or absent in a particular image. Taking into account that normal images constitute the majority of the dataset, using cross-entropy loss aids maintaining training stability, as it evaluates each pixel independently of PE existence or size.

The cross-entropy loss for a single image can be defined as follows:

$$L_{\mathrm{CE}} = -\frac{1}{N}\sum_{i=1}^{N}\sum_{c=1}^{C} y_{i,c}\, (p_{i,c}) \tag{3}$$

where $N$ denotes the number of pixels in an image, $C$ is the total number of classes, $y_{i,c} \in \{0,1\}$ represents the ground truth label for pixel $i$ and class $c$ and $p_{i,c} \in [0,1]$ is the predicted probability of pixel $i$ being assigned to class $c$.

The feature matching loss leverages both labeled ($x^l$) and unlabeled ($x^u$) images and is designed to minimize the discrepancy between feature representations of predicted segmentation maps ($S(x^u)$) and ground truth masks ($y^l$). In other words, this loss encourages the generator to generate predictions, which are similar to those of the real annotations. Feature matching loss is defined as follows:

$$L_{\mathrm{FM}} = \left\| \mathrm{E}_{(x^l, y^l)\sim D^l}[D_k(y^l \oplus x^l)] - \mathrm{E}_{x^u\sim D^u}[D_k(S(x^u) \oplus x^u)] \right\|, \tag{4}$$

where $D_k(\cdot)$ refers to the feature maps extracted from the intermediate $k$-th layer of the discriminator.

The self-training loss only utilizes the unlabeled images ($x^u$). Specifically, it leverages the artificial ground truth masks, generated by the network that successfully 'fool' the discriminator. These high-confidence predictions are then treated as pseudo-ground truth ($y^*$) for their respective unlabeled input scans and are incorporated into the training process. The self-training loss is defined as follows:

$$L_{\mathrm{ST}} = \begin{cases} -\sum_{h,w,c} y^* \log \mathrm{S}(x^u) & , \text{if } D(S(x^u)) \geq t \\ 0 & , \text{otherwise} \end{cases} \tag{5}$$

Taking into account that the discriminator outputs a probability score indicating whether a sample belongs to the real distribution, only those predicted segmentation masks with discriminator scores above a predefined threshold $t$ are selected and used in the loss computation. This ensures that only high-confidence pseudo-labels contribute to training.

The generator's total loss $Ls$ is computed as a weighted sum of the 4 previously defined losses and is formally expressed as follows:

$$L_s = \lambda_{Dice} L_{\mathrm{Dice}} + \lambda_{CE} L_{\mathrm{CE}} + \lambda_{\mathrm{FM}} L_{\mathrm{FM}} + \lambda_{\mathrm{ST}} L_{\mathrm{st}} \tag{6}$$

where $\lambda_{Dice}$, $\lambda_{CE}$, $\lambda_{\mathrm{FM}}$, $\lambda_{\mathrm{ST}}$ are the weights of $L_{\mathrm{Dice}}$, $L_{\mathrm{CE}}$, $L_{\mathrm{FM}}$ and $L_{st}$ respectively, adjusting the contribution of each loss term.

### *B. Discriminator*

The discriminator network is trained to distinguish artificial from real ground truth masks. It is trained using the original GAN loss proposed in [21] as shown in Eq. 7. It takes as input pairs comprising the CTPA image and its accompanying ground truth mask and outputs a probability score indicating whether the ground truth mask is artificial or real. The loss $L_D$ used to train the discriminator is defined as:

$$L_{\mathrm{D}} = \mathrm{E}_{(x^l, y^l)\sim D^l}[\log D(y^l \oplus x^l)] + \mathrm{E}_{x^u\sim D^u}[\log(1 - D(S(x^u)) \oplus x^u)], \tag{7}$$

where $\oplus$ denotes concatenation along the channel dimension. By improving its ability to differentiate artificial from real samples, the discriminator provides feedback, aiding the generator to refine its output.

## III. Experimental Evaluation

This section provides details on the publicly available dataset used and the experimental setup, as well as the results of various experiments and an ablation study to evaluate the contribution of each element of the proposed method to the segmentation accuracy obtained.

### *A. Dataset*

We evaluate the proposed method using a publicly available dataset called FUMPE [22], which is currently the only publicly available dataset with pixel-wise annotations for PE segmentation. FUMPE contains 8792 CTPA slices, accompanied by pixel-wise ground truth masks. The proposed dataset totally includes 3438 PE-ROIs. Most of these ROIs (67%) are located in the peripheral arteries, making this dataset particularly challenging. We use 70% of the available samples for training and 30% for testing.

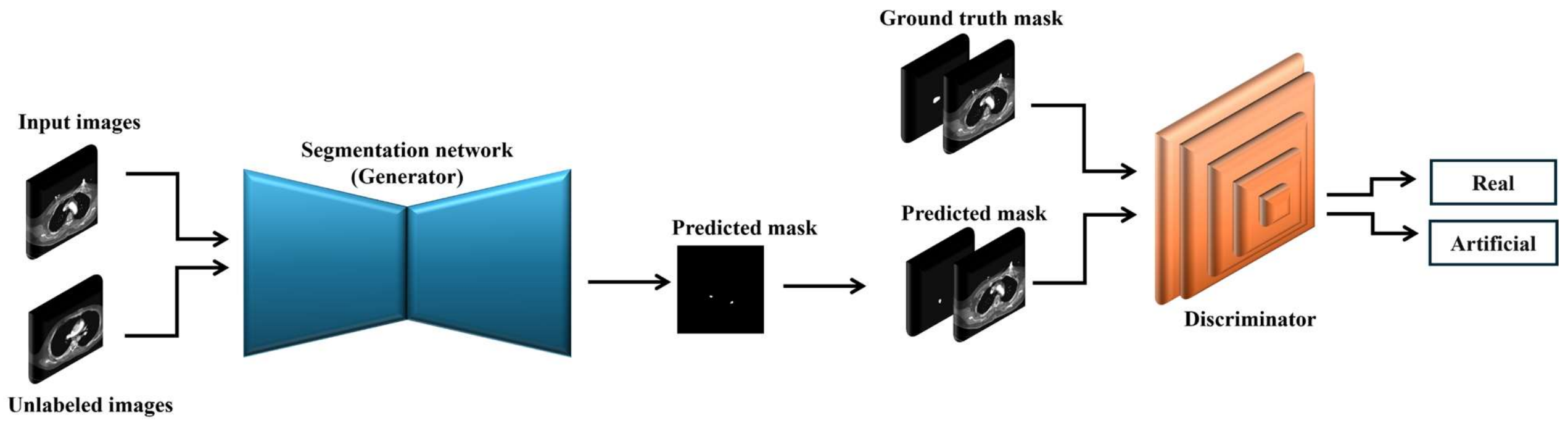

Fig. 1. Overview of the proposed S4GAN-PE method.

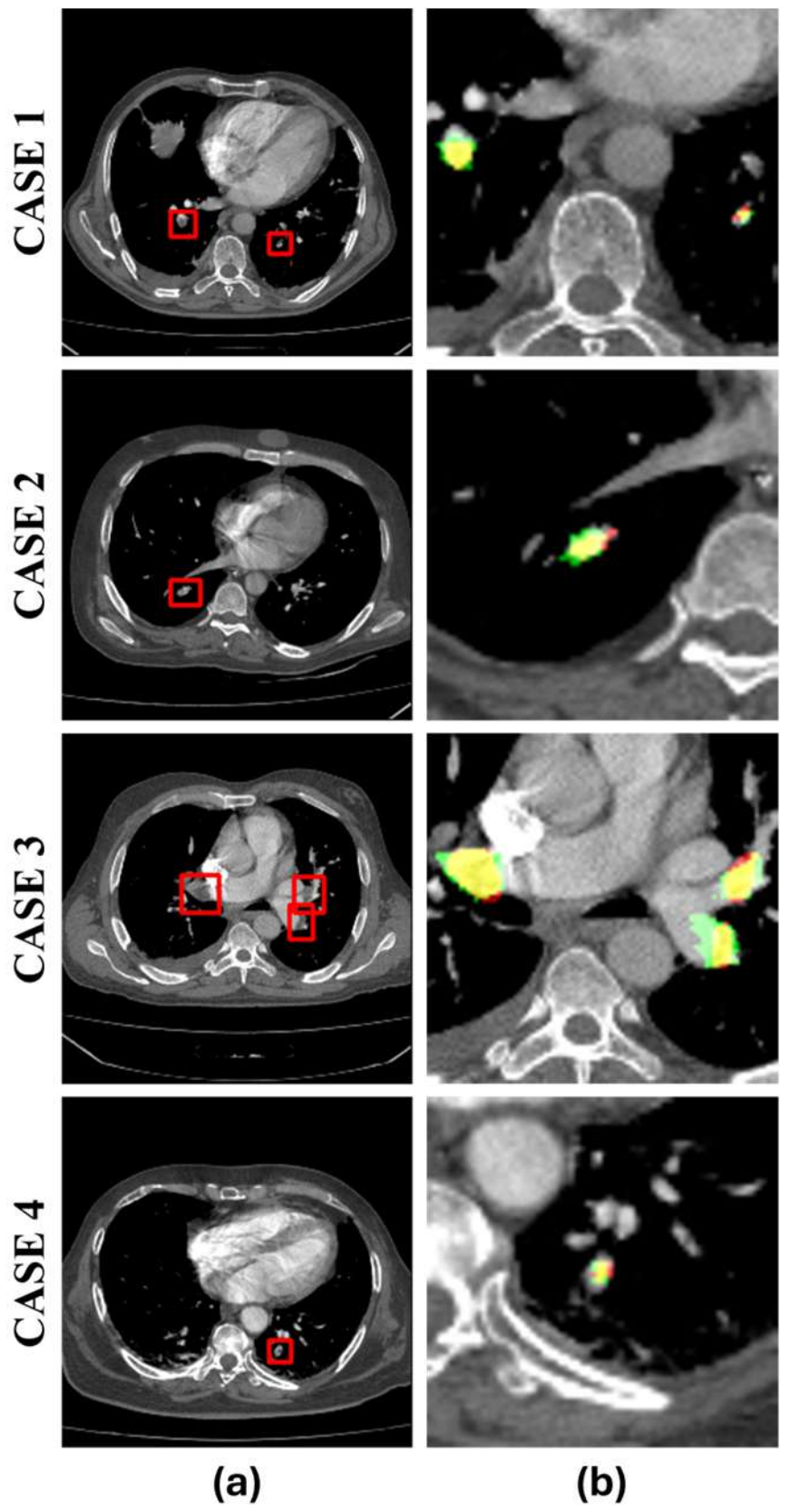

Fig 2. Evaluation results on the test set of the FUMPE dataset. (a) input CTPA slices (PE regions marked by solid red boxes), (b) overlaid visualization of the ground truth mask (red) and the predicted segmentation mask (green). Overlapped regions are shown in yellow.

## B. *Experimental Setup*

For the semi-supervised setup, 30% of the training set is used as labeled data, while the remaining 70% is used as an unlabeled data pool. More extreme low-label regimes (e.g., 5%) yield very few PE-positive slices, leading to unstable training and unreliable evaluation. As the segmentation backbone, we utilize SCUNet++, a fully supervised architecture, tailored for PE, which integrates the strengths of UNet++, multiple fusion dense skip connections, SWIN-Transformer attention mechanism, and SWIN-UNet [12].

Following the original discriminator of the s4GAN implementation, the discriminator is composed of four convolutional layers with kernel size 4×4 and channel dimensions of 64, 128, 256, 512. Each convolutional layer is followed by Leaky ReLU activation, with a negative slope of 0.2, and a dropout layer with probability of 0.5, to mitigate overfitting and promote stable training. Moreover, the training of the generator uses Stochastic Gradient Descent (SGD), with a learning rate of $2.5 \times 10^{-4}$, while the discriminator is trained using Adam, with a learning rate of $10^{-4}$.

TABLE I. COMPARISON WITH STATE-OF-THE-ART ON FUMPE

| Method | Dice score | HD95 |
|---|---|---|
| SCUNet++ [12] | 83.47 | 3.83 |
| [17] | 56.11 | - |
| Proposed method | **88.75** | **1.74** |

With respect to hybrid Dice loss and cross-entropy loss, we assign weights of 0.6 and 0.4 to each component, respectively. The weights of the different loss components are empirically determined as $\lambda_{FM}$=0.1 , $\lambda_{ST}$=1 and the confidence threshold $t$ used to filter reliable predictions is set to 0.6. We use a batch size of 12 for all experiments. All networks were trained and evaluated on a workstation equipped with an NVIDIA GeForce RTX 3090 GPU.

## C. *Results*

In this Section, we present a comprehensive qualitative and quantitative analysis of the proposed method, including comparisons with two baseline methods: the fully-supervised SCUNet++ [12] and the semi-supervised method introduced in [17], which, to the best of our knowledge, is the only semi-supervised method for CTPA image segmentation.

As illustrated in Table I the proposed method achieves the highest Dice score, surpassing both baseline methods. Although the proposed method is semi-supervised, utilizing part of the available labeled samples for training, it outperforms the fully supervised SCUNet++, with respect to both metrics. It should be noted that these results are obtained despite the class imbalance at both the slice level (rare positive slices) and pixel-level (PE occupying a tiny fraction of the image).

Figure 2 provides example segmentation results of the proposed method on CTPA scans of the FUMPE dataset. By visual comparison with the ground truth masks, it can be observed that all PEs are accurately localized and delineated. This includes cases of small emboli (CASE 2, CASE 4 in Fig.2), which present a pronounced class imbalance at the pixel level, as well as cases of CTPA slices with multiple emboli (CASE 1, CASE 3 in Fig.2).

*D. Ablation Study*

Aiming to evaluate the contribution of semi-supervised learning in PE segmentation, we conducted a series of ablation experiments focusing on three key factors: the amount of unlabeled data, the quantity of labeled data and the network architecture.

*1) Effect of the number of unlabeled data:* We varied the number of unlabeled training samples, while keeping the number of labeled samples fixed (at 30% of the total labeled samples in the FUMPE dataset, represented by ratio=0.3), in order to evaluate the effect of the amount of unlabeled data on segmentation performance. The results presented in Table II demonstrate that increasing the amount of unlabeled data, consistently improves the segmentation performance. As expected, the segmentation performance reaches its maximum (Dice score and IoU equal to 89.29% and 86.18%, respectively), when all available unlabeled data were utilized.

*2) Effect of the number of labeled data:* We vary the number of labeled training samples, in order to evaluate the effect of the amount of labeled data on the segmentation performance. As shown in Table III, increasing the number of labeled training samples (expressed as a ratio of the total labeled samples in the FUMPE dataset), generally leads to higher segmentation accuracy. This is expected considering the stronger supervision. However, when examining the trade-off between annotation cost and model performance, it can be observed that the proposed semi-supervised method achieves competitive segmentation results, even with a limited amount of labeled samples for training. This highlights its ability to learn effectively from unlabeled samples. For instance, by using only 30% (ratio=0.3) of the labeled samples and treating the remaining 70% as unlabeled, the proposed method achieves a Dice score of 85.76%, which is comparable to the Dice scores obtained using more than 50% of the labeled samples. Furthermore, Table III shows that precision consistently exceeds recall across all labeling ratios, although the two metrics remain well-balanced.

*3) Architecture of segmentation network:* Aiming to evaluate the impact of backbone network architecture on segmentation accuracy in terms of the IoU and Dice coefficient, we experimented with two different architectures: (i) DeepLab-v2 [23], an architecture widely used in natural image segmentation, and (ii) SCUNet++, a fully supervised network architecture, specifically designed to cope with CTPA image segmentation and PE delineation. The results of these experiments are presented in Table III, where the segmentation accuracy of each architecture is reported for different labeled data ratios. It can be observed that DeepLab-v2 follows the same overall trend as SCUNet++: as the ratio of labeled samples increases, segmentation accuracy also increases. However, SCUNet++ consistently outperforms DeepLab-v2 across all labeled sample ratios, with respect to both Dice and IoU metrics.

TABLE II. PERFORMANCE ACROSS VARYING UNLABELED DATA RATIOS

| ratio | IoU | Dice | Recall | Precision |
|---|---|---|---|---|
| Fully-supervised | 0.8235 | 0.8556 | 0.8544 | 0.8707 |
| 0.1 | 0.8453 | 0.8739 | 0.8634 | 0.9021 |
| 0.3 | 0.8528 | 0.8637 | 0.8607 | 0.8811 |
| 0.5 | 0.8601 | 0.8842 | 0.8848 | 0.8969 |
| 0.8 | 0.8533 | 0.8853 | **0.8970** | 0.8894 |
| 1.0 | **0.8618** | **0.8875** | 0.8949 | **0.9025** |

## IV. CONCLUSIONS

In this work, we proposed a semi-supervised adversarial learning method for PE boundary extraction in CTPA scans. The proposed method leverages a set of labeled CTPA scans, as well as a larger pool of unlabeled scans by employing semi-supervised learning in the form of adversarial training. In addition, Dice loss and cross-entropy loss are employed to tackle the highly imbalanced nature of CTPA scans and the small size of PEs. Feature matching loss encourages the generation of anatomically plausible artificial ground truth masks, indistinguishable from real ground truth, and self-training loss reinforces learning from highly-confident pseudo-labeled samples. The experimental evaluation leads to the following conclusions:

*i.* A fully-supervised network like SCUNet++ can achieve comparable, or even superior results in the CTPA image segmentation task, when integrated within a semi-supervised GAN-based framework. As shown in Section III.C, even with only 30% of the total labeled samples in the FUMPE dataset, the proposed method accurately localizes and delineates PEs, surpassing the segmentation accuracy of the fully-supervised SCUNet++, as well as of the method proposed in [17].

*ii.* Most PEs are located in peripheral veins. Accordingly, they are very small and difficult to detect. The high IoU and Dice scores achieved by the proposed method demonstrate its strong capability to accurately extract even these challenging emboli, as illustrated in Fig 2.

*iii.* The proposed method effectively leverages not only the supervised signal from labeled samples but also crucial information from unlabeled samples, and accurately localizes and delineates PEs, regardless of their number or size. Ultimately, the proposed method relies less on costly and labor-intensive annotations, while it maintains high segmentation accuracy.

Future work includes the application of the proposed semi-supervised method to other challenging medical image segmentation tasks, which are characterized by limited annotations and class imbalance. Additionally, to further optimize annotation efficiency, active learning can be integrated in the proposed method.

TABLE III. RESULTS OF DIFFERENT SEGMENTATION MODELS AS GENERATORS ACROSS VARIOUS LABELED DATA RATIOS.

| | IoU | | Dice | | Recall | | Precision | |
|---|---|---|---|---|---|---|---|---|
| **ratio** | **DeepLab-v2** | **SCUNet++** | **DeepLab-v2** | **SCUNet++** | **DeepLab-v2** | **SCUNet++** | **DeepLab-v2** | **SCUNet++** |
| 0.1 | 0.6365 | **0.7921** | 0.6671 | **0.7921** | 0.7184 | **0.8181** | 0.7184 | **0.8355** |
| 0.3 | 0.6837 | **0.8576** | 0.7164 | **0.8875** | 0.7121 | **0.8903** | 0.7344 | **0.8979** |
| 0.5 | 0.5262 | **0.8638** | 0.5615 | **0.8936** | 0.5648 | **0.8998** | 0.5713 | **0.8988** |
| 0.8 | 0.6936 | **0.8797** | 0.7259 | **0.9105** | 0.7264 | **0.9098** | 0.7337 | **0.9211** |